\documentclass{article}
\usepackage[margin=1.3in]{geometry}
\usepackage[portuges]{babel} 
\usepackage[utf8x]{inputenc}  

\usepackage{fullname_pt}

\title{Methodology and Results for the Competition on Semantic Similarity Evaluation and Entailment Recognition for PROPOR 2016}

\author{
  Luciano Barbosa\\
  \texttt{IBM Research}
  \and 
  Paulo R. Cavalin\\
  \texttt{IBM Research}
  \and
  Victor Guimarães\\
  \texttt{IBM Research}
  \and 
  Matthias Kormaksson\\
  \texttt{IBM Research}
}

\begin{document}
\maketitle

\begin{abstract}
   In this paper, we present the methodology and the results obtained by our teams, dubbed
   Blue Man Group, in the ASSIN (from the Portuguese {\it Avaliação de Similaridade Semântica 
   e Inferência Textual}) competition, held at PROPOR 2016\footnote{International Conference 
   on the Computational Processing of the Portuguese Language - http://propor2016.di.fc.ul.pt/}. 
   
   Our team's strategy consisted of evaluating methods based on semantic word vectors, 
   following two distinct directions: 1) to make use of low-dimensional, compact, feature sets, 
   and 2) deep learning-based strategies dealing with high-dimensional feature vectors. 
   Evaluation results demonstrated that the first strategy was more promising, so that the
   results from the second strategy have been discarded.
   
   As a result, by considering the best run of each of the six teams, we have been able to 
   achieve the best accuracy and F1 values in entailment recognition, in the Brazilian 
   Portuguese set, and the best F1 score overall. In the semantic similarity task, our
   team was ranked second in the Brazilian Portuguese set, and third considering both
   sets.
\end{abstract}


\section{Introduction}
In this work, we present the methodology and results obtained by our team, dubbed Blueman
group, in the {\it Avaliação de Similaridade e Inferência Textual} (ASSIN) competition, 
jointly
held with the International Conference on the Computational Processing of Portuguese (PROPOR) 
2016. 

The ASSIN competition assigned two 
tasks to participants: evaluation of semantic similarity, and entailment recognition. Given 
sentences $s_1$ and $s_2$, the first task consists of 
providing a score ranging from 1 to 5, representing the strength of the semantic relationship between
$s_1$ and $s_2$. The second task involves determining whether $s_1$ entails $s_2$
(a sentence $s_1$ entails another sentence $s_2$ if, after reading both and knowing that $s_1$ is true, a person concludes that $s_2$ must also be true). Given these two tasks, researchers are invited to form teams and 
participate in the competition by developing systems that solve either or both of them, by 
making use of labeled data provided by the organization of the competition, and submit their 
results on a blind test data, the accuracy of which is used to rank the teams and define the 
winners. It is worth mentioning that sets with text in Portuguese from both Brazil and 
Portugal were available, i.e. PT-BR and PT-PT, and teams could choose to submit results 
for either or both sets.

Our team (Blueman group) focused on word vectors-based approaches to solve both tasks (see details in section \ref{sec:methodology}).
By considering word vectors created with the entire Portuguese Wikipedia, we have followed
two distinct directions. In the first, we implement a state-of-the-art feature set, proposed
in \cite{Kenter2015}, to train both support vector regression/classification models and
Lasso regression. In the second direction, we exploit deep-learning setups of siamese
neural networks. Preliminary evaluations on the training and trial data sets demonstrated that
the first direction was more promising, and we have decided to demonstrate the results of
that methodology only.

In total, six teams participated in the competition. By considering the best run of
each team, our system worked best in the entailment recognition task, ranking first in 
both accuracy and F1 for the PT-BR set, while ranking second in accuracy and first in F1 overall.
In the semantic similarity evaluation, our best results were ranked second in both Pearson correlation
and Mean Squared Error (MSE) for the PT-BR set, while ranking second in Pearson and third in MSE
overall. For the PT-PT set, the system performed better for entailment recognition,
achieving the second best F1 score, while achieving only the 4th place in semantic
similarity.

In the remainder of this document we present details on how our system was developed and evaluated.

\section{ASSIN Competition}
The ASSIN competition, a.k.a {\it Avaliação de Similaridade Semântica e Inferência Textual}, consists
of an evaluation forum for two NLP-related tasks, i.e. semantic similarity and textual entailment recognition, 
where registered participants (or teams) could develop systems and submitted their results on the data provided 
by the organizing committee. A large dataset containing pairs of sentences, in both Portugal's and Brazil's
variants of Portuguese, has been created to allow participants to both develop and evaluate the systems.
And the participants could submit results to either or both task, and also either or both variations
of Portuguese. Then, the teams would be ranked by the results of their systems on the evaluation
dataset, namely test set. Both the metrics and the sets, as well as task, are explained in detail, 
as follows.

The ASSIN dataset, containing a total of 10,000 pairs of sentences, can be divided in the following 
subsets. The Brazilian training set contains 3,000 labelled pairs of sentences collected from Google News,
from Brazilian sources. The Portuguese training set also contains 3,000 labelled pairs of sentences collected from
Google News, but from Portuguese sources. And the Brazilian and Portuguese blind test sets, contain 2,000
unlabeled pairs of sentences each, from the same sources. It is worth mentioning that the labels of the test
sets have been released to the participants only after they had submitted their results.

For the first task, i.e. semantic similarity, the semantic relatedness is measure in a scale from 1 to 5, 
where 1 stands for completely different sentences, and 5 sentences that means essentially the same thing. 
The scales in between are gradual variations of these two concepts. In the light of this, this task consists
of building a model which, given the pair of sentences $p(i) = \{s_1(i), s_2(i)\}$, containing sentence
$s_1(i)$ and sentence $s_2(i)$, predicts the semantic similarity score $y(i)$. Given the manually-labeled
similarity scores $x(i)$, systems are evaluated by means of the Pearson correlation between the set containing all
$x(i)$ and $y(i)$, for $i > 0$, and the Mean Squared Error (MSE).

The second task, i.e. recognizing textual entailment (RTE), consists of determining whether
the meaning of the hypothesis is entailed from the text \cite{RTE2016}. That is, suppose
$s_1$ is the text and $s_2$ is the hypothesis, $s_1$ entails $s_2$ if, after reading
both and knowing that $s_1$ is true, a person concluded that $s_2$ must also be true. 
Given that the dataset provided by ASSIN also distiguishes bidirectional entailment cases,
or paraphrases, the pair of sentences $s_1$ and $s_2$ must be classified into one of the
following classes: {\it entailment}, {\it paraphrase}, and {\it no relation}. Given the 
ground-truth labels, systems are measured by means of accuracy and F1 score.

More details regarding ASSIN are available at \cite{ASSIN2016}.

\section{Methodology}
\label{sec:methodology}
As already mentioned, the strategy employed by our team consisted in evaluating word vector-based
approaches, where the word vectors represent the semantic meaning of words (see Section~\ref{sec:word_vectors}). 
As a result, two distinct directions have been followed. The first, presented in Section~\ref{sec:strategy1},
consists of implementing a state-of-the-art
feature set for representing the similarity relatedness of pairs of sentences, and using
regression models such as support vector regression (SVR) for semantic similarity evaluation,
and support vector machines (SVM) for entailment recognition. And the second, in Section~\ref{sec:strategy2}, 
exploits deep-learning siamese neural networks, with the goal of learning better representation from
raw data, i.e. the word vectors of the pair of sentences.

\subsection{Word vectors}
\label{sec:word_vectors}
Word vectors (or word embeddings) have been successfully used over the past years to learn useful
word representations, encoding the semantic meaning of words by means of continuous vectors 
\cite{Collobert2011}. In other words, even if two words are lexically written in two very
distinct ways, if these two words present similar semantic meaning, their corresponding
word vectors should be very similar. These vectors make it possible not only to create NLP 
methods that rely more on the semantic meaning of the words than on their lexical form,
but to take advantage of large corpora of text since the learning of word vectors can be
done in an unsupervised fashion. 

The learning of word vectors is done in the following way. Given a large corpus of text,
word vectors are learned by considering the distributional frequency of words. That is, 
given a word and its preceding and subsequent words in a sentence, a machine learning model
such as a neural network can be learned by using the neighbouring words are input, and the
central word as output.

In this work, word vectors have been created with the {\it word2vec} 
tool\footnote{http://code.google.com/archive/p/word2vec/}, using the entire
Portuguese Wikipedia as input. This set used contains a total of 636,597 lines of texts,
with 229,658,430 word occurrences, and a vocabulary of size 540,638.
The {\it word2vec} tool was setup with: skip n-grams model; word vector size equals to 300; 
maximum skip length between words set to 5; 10 negative samples; hierarchical softmax not used;
threshold of occurrence of words set to 10e-5; and 15 training iterations.

\subsection{Strategy 1: Kenter's features}
\label{sec:strategy1}

\subsubsection{Feature set}
The feature set proposed in \cite{Kenter2015}, consists of extracting a single feature vector,
denoted $\bar{x}_i = x_{i1},\dots,x_{iK}$, to encode the semantic similarity from the pair of sentences
$s_{i,1}$ and $s_{i,2}$. In this work, we propose the use of such feature set for both tasks in the 
competition, i.e. semantic similarity evaluation and entailment recognition.

Given the sets word vectors $\Omega_{i,1}$ and $\Omega_{i,2}$, computed from sentences $s_{i,1}$ and 
$s_{i,2}$, the feature set is composed of two types of features. 1) semantic networks; and 2) text-level features.

In short, semantic networks consist of building a network considering the distances of pairs of terms
that appear in $s_{i,1}$ and $s_{i,2}$. In this case, two types of networks are built. The first, namely
Saliency-weighted Semantic Network, combines both similarity and inverse document frequency (IDF) to create
the links between the nodes, by considering, for each term in $s_{i,1}$, the most similar term in $s_{i,2}$,
i.e. the terms with the most similar word vector. The second type of network, referred to as Unweighted
Semantic Network, in contrast, does not rely on IDF, and two different unweighted networks are derived from
this. One contains the distance of the word vector of all terms to the other ones, and the other only the
maximum distance. In the end, the information in these networks is used to create histograms, which are concatenated
to compose a single feature vector. 

Text-level features are defined in two ways: 1) distance between word vectors, where both the cosine and
Euclidean distances are computed between the mean word vectors of $s_{i,1}$ and $s_{i,2}$; and 2) bins of
dimensions, where a histogram is computed from the real values presented in the mean word vectors of the
pair of sentences.

The boundaries for the aforementioned histograms have been defined in the following way. For the features
calculated from the saliency-weighted semantic network, the values are 0-.15, .15-.4, and .4-$\infty$.
For the unweighted semantic network, the values are -1-.45, .45-.8, and .8-$\infty$. And for the bins 
of dimension, the values are $-\infty$-.001, .001-.01, .01-.02, and .02-$\infty$. Details on how these
boundaries have been defined, along with values for other parameters, can be found in 
\cite{Kenter2015}.

The resulting feature set consists of a 15-position vector, based on: 3 features from histogram of
saliency-weighted semantic networks, 2 $\times$ 3 from the histograms from the unweighted semantic networks,
2 from the distances of the mean word vectors, and 4 from the bins of dimension. It is worth mentioning
that these 15 features can be replicate other set of word vectors, but in this work, we consider only
the word vectors described in Section~\ref{sec:word_vectors}.

\subsubsection{Support Vector Regression and Support Vector Machines}
Support vector machines (SVM), and their corresponding method for regression problems, i.e. Support Vector
Regression (SVR), have become popular in the past years given the good performance in a high number of
tasks [citation]. SVM and SVR employ the following idea: input vectors, denoted $x_{i1},\dots,x_{iK}$,
are non-linearly  mapped to a very  high-dimension  feature space. In this feature space, a linear
decision surface is constructed, in order to predict the class value $y_i \in [-1,1]$, in the
case of classification, or the target real value $y_i$, in the case of regression. Special properties of the 
decision  surface ensures high  generalization  
ability of the learning  machine \cite{Cortes1995}.

For this work, both SVR and SVM have been implement with the {\it Scikit Learn} 
library\footnote{http://scikit-learn.org}.
For both methods, we used the Gaussian kernel after a few preliminary experiments. And the 
configuration parameters of both have been setup by means of a grid search with five-fold
cross validation.

\subsubsection{Lasso}

Let $y_i$ denote the response and let $x_{i1},\dots,x_{iK}$ denote the $K$ features calculated for each observation $i$. We considered the following regression model:
$$
y_i = \beta_0 + \sum_{k=1}^K \beta_k x_{ik} + \sum_{\ell \neq k} \alpha_{\ell k} x_{i\ell}x_{i k} + \varepsilon_i,
$$ 
where $\varepsilon_i$ denotes the error associated with observation $i$. The above model is linear in the features and includes all possible two-way interactions, $x_{i\ell}x_{ik}$, between pairs of features. Let $\theta$ denote the set of all parameters $(\beta_k)_k$ and $(\alpha_{\ell k })_{\ell k}$. By correctly specifying a design matrix $X$ (whose columns are the features and corresponding two-way interactions) we may formulate the above regression in a more simple matrix notation:
$$
y = X \theta + \varepsilon,
$$
where $y$ and $\varepsilon$ are the response and error vectors respectively. 

Note that if we were to estimate the above model using the method of least squares we could easily have problems with over-fitting due to the large amount of parameters to be estimated: 
$$
n_{param} = K + 1 + \frac{(K-1)\cdot K}{2} \sim O(K^2).
$$
Lasso regression is designed to tackle this potential problem of over-fitting and falls into a class of models called regularized regression. By applying least squares with an additional $L_1$-constraint on the parameters, $\| \theta \|_1 = \sum_k |\theta_k| \leq C$, for some $C > 0$, we are able to guard against over-fitting. This method has an advantage in that it serves as a method for variable selection as well, since the $L_1$-penalty effectively forces some of the parameter estimates to be exactly equal to $0$.

\subsection{Strategy 2: Siamese Networks}
\label{sec:strategy2}

Siamese networks~\cite{} have been widely used in image and text processing to learn a similarity metric from data.
For the specific task proposed on ASSIN, we use siamese networks to learn the similarity between two sentences in Portuguese. Essentially, given a pair of sentences, a siamese network projects each sentence in a new representation space using, for instance, convolutional or recurrent networks. The parameters W of each sentence branch are shared.
These representations are then given as input to a pre-defined similarity metric such as cosine or euclidean that calculates the similarity between the two representations. During training, the network learns the values of W that minimize a given loss function. In our experiments, we use Mean Squared Error as the loss function. The error is the difference between the true similarity value given in the training data and the predicted one. From this framework, we tried different configurations. For instance, to project the sentences we tried convolutional and recurrent networks, and as similarity metrics cosine and dot product. In the experimental evaluation, we present the siamese networks that obtained the best results over the test set.

lstm 0.47, 0.04

\section{Evaluation Results}
In this section, we discuss the results obtained with the methods described in 
Section~\ref{sec:methodology}. For such an evaluation, we consider the Trial dataset
as test set, and both PT-BR and PT-PT training sets. Note that we have removed from
PT-BR the samples that also appear in Trial.

A comparison of the results for each method is presented in 
Table~\ref{tab:evaluation_results}. In this case, the best results have been achieved 
with Kenter's features with either SVRs or Lasso for semantic similarity evaluation,
and SVMs for entailment recognition. With SVR, Pearson correlation of 0.51, 0.49, and 0.50
have been reached for PT-BR, PT-PT, and Overall sets, respectively. In the entailment
recognition task, F1 scores of 0.45, 0.50, and 0.51, have been achieved on the same
sets, respectively. In addition, we observe that with Lasso, the results are very similar
to those of SVR.

\begin{table*}[h]
\centering
\caption{Evaluation results, considering Trial as test set.}
\label{tab:evaluation_results}
\begin{tabular}{|c|c|c|}
\hline
{\bf Configuration} & {\bf Similarity} & {\bf Entailment}	\\
\hline
Baseline: Bag of Words Overall & 0.47 &  \\
\hline
Kenter's features - SVR(M) PT-BR & 0.51 & 79.60/0.45\\
Kenter's features - SVR(M) PT-PT  & 0.49 & 74.20/0.50\\
Kenter's features - SVR(M) Overall & 0.50 & 77.00/0.51 \\
Kenter's features - Lasso PT-BR	&  & \\
Kenter's features - Lasso PT-PT  &  & \\
Kenter's features - Lasso Overall & & \\
\hline
LSTM várias camadas + reg L2 & 0.26	& \\ 
LSTM várias camadas + reg L2 + features & 0.23 & \\
CNN & 0.13 & \\
LSTM várias camadas + reg L2 + features + Full Data & 0.49 & \\
CNN + Cos PT-BR & 0.35 & \\
LSTM + Cos PT-BR & 0.41 & \\
LSTM + Cos Overall & 0.38 & \\
LSTM + Concat + Kenter's features PT-BR & 0.33 & \\
LSTM + Cos + Kenter's features PT-BR & 0.33 & \\
LSTM + Dot + Kenter's features PT-BR & 0.29 & \\
LSTM (Cos) + BOW (Cos) + Kenter's features PT-BR & 0.39 & \\
CNN (Cos) + BOW (Cos) + Kenter's features PT-BR & 0.40 & \\
BOW (Cos) + BR  & 0.34 & \\
LSTM (Cos) + CNN (Cos) + BOW (Cos) + Kenter's features PT-BR & 0.38 & \\
\hline
\end{tabular}
\end{table*}

The second strategy, making use of Siamese networks, has not achieved good results. The best
results with this method were 0.11 points below that from strategy 1. For this reason,
we decided to submit the results only with Kenter's features, one run with SVR and
another run with Lasso for semantic similarity, and one run with SVM in entailment 
recognition.

\section{Competition Results}
In this section we discuss the results of our methods in the blind test data, and how it
compared with the other competitors.

In total, six teams participated in the competition. In addition our team, only two other 
teams submitted results for both tasks and both PT-BR and PT-PT sets. From the remaining
three teams, two have focused only on the semantic similarity task, considering both sets,
and the other one only on the PT-PT set, for both similarity and entailment recognition
tasks. 

The best result of each team\footnote{Each team was allowed to submit up to three different
runs}, i.e. the best run, is listed in Table~\ref{tab:teams_results}, and the ranking of
each team, considering only the best run, is presented in Table~\ref{tab:teams_ranking}. Considering
only the best run of each team, we have managed to achieve very good results with the PT-BR
set and Overall, being far from the first place only in the PT-PT set. With PT-BR, we ranked
first in both accuracy and F1 metrics for entailment recognition, and second best in semantic similarity
evaluation. Besides the good results, it was surprising that Kenter's features performed better
in entailment recognition than semantic similarity evaluation, since the feature set has been
originally proposed for the latter task. Overall, we ranked first in entailment recognition in
F1, and second in accuracy. In semantic similarity, our team presented the second best Person
correlation values, and the third best MSE value. In the PT-BR set, we have been able to be ranked
second in F1 for the entailment recognition, and third in accuracy. But for semantic similarity, only
the fourth place (tied with another team) has been reached. 

\begin{table*}[h]
\centering
\caption{Best results of each team in the competition.}
\label{tab:teams_results}
\begin{tabular}{|p{2cm}|c|c|c|c|c|c|c|c|c|c|c|c|}
\hline

    & \multicolumn{4}{|c|}{\bf PT-BR} & \multicolumn{4}{|c|}{\bf PT-PT} & \multicolumn{4}{|c|}{\bf Overall}	\\
       \hline
    & \multicolumn{2}{|c|}{\bf Sim} & \multicolumn{2}{|c|}{\bf RTE} & 
      \multicolumn{2}{|c|}{\bf Sim} & \multicolumn{2}{|c|}{\bf RTE} & 
       \multicolumn{2}{|c|}{\bf Sim} & \multicolumn{2}{|c|}{\bf RTE} \\
       \hline
{\bf Team} & {\bf P} & {\bf MSE} & {\bf Acc} & {\bf F1} & {\bf P} & {\bf MSE} & {\bf Acc} & {\bf F1} & {\bf P} & {\bf MSE} & {\bf Acc} & {\bf F1} \\
\hline
\hline
Solo Queue & 0.70 & 0.38 & - & - & 0.70 & 0.66 & - & - & 0.68 & 0.52 & - & - \\
\hline
Reciclagem & 0.59 & 1.31 & 79.05 & 0.39 & 0.54 & 1.10 & 73.10 & 0.43 & 0.54 & 1.23 & 75.58 & 0.40 \\
\hline
ASAPP & 0.65 & 0.44 & 81.65 & 0.47 & 0.68 & 0.70 & 78.90 & 0.58 & 0.65 & 0.58 & 80.23 & 0.54 \\
\hline
LEC-UNIFOR & 0.62 & 0.47 & - & - & 0.64 & 0.72 & - & - & 0.62 & 0.59 & - & - \\
\hline
L2F/INESC-ID & - & - & - & - & 0.73 & 0.61 & 83.85 & 0.70 & - & - & - & - \\
\hline
\hline
{\bf Blue Man Group} & 0.65 & 0.44 & 81.65 & 0.52 & 0.64 & 0.72 & 77.60 & 0.61 & 0.63 & 0.59 & 79.62 & 0.58 \\
\hline
\end{tabular}
\end{table*}

One observation that is worth mentioning, is that in some tasks or sets that teams that achieved the best
results were those that focused only in one task or set. For instance, the {\it Solo Queue} team submitted
results only for semantic similarity, and they won the task for PT-BR and Overall, and ranked second for
PT-PT. The {\it L2F/INESC-ID} team, on the other hand, submitted results only for PT-PT, for both tasks,
and they won both. In our case, we submitted a single method, we almost no difference from one set to another, 
or from one task to another. As lessons learned, in a future competition, we believe we shall invest more
on fine tuning the algorithms to specific tasks and sets.

\begin{table*}[h]
\centering
\caption{Teams ranking considering the best run.}
\label{tab:teams_ranking}
\begin{tabular}{|p{2cm}|c|c|c|c|c|c|c|c|c|c|c|c|}
\hline

    & \multicolumn{4}{|c|}{\bf PT-BR} & \multicolumn{4}{|c|}{\bf PT-PT} & \multicolumn{4}{|c|}{\bf Overall}	\\
       \hline
    & \multicolumn{2}{|c|}{\bf Sim} & \multicolumn{2}{|c|}{\bf RTE} & 
      \multicolumn{2}{|c|}{\bf Sim} & \multicolumn{2}{|c|}{\bf RTE} & 
       \multicolumn{2}{|c|}{\bf Sim} & \multicolumn{2}{|c|}{\bf RTE} \\
       \hline
{\bf Team} & {\bf P} & {\bf MSE} & {\bf Acc} & {\bf F1} & {\bf P} & {\bf MSE} & {\bf Acc} & {\bf F1} & {\bf P} & {\bf MSE} & {\bf Acc} & {\bf F1} \\
\hline
\hline
Solo Queue          & 1st & 1st & - & - & 2nd & 2nd & - & - & 1st & 1st & - & - \\
\hline
Reciclagem          & 5th & 5th & 3rd & 3rd & 6th & 6th & 4th & 4th & 5th & 5th & 3rd & 3rd \\
\hline
ASAPP               & 2nd & 2nd & 1st & 2nd & 3rd & 3rd & 2nd & 3rd & 2nd & 2nd & 1st & 2nd \\
\hline
LEC-UNIFOR           & 4th & 4th & - & - & 4th & 4th & - & - & 4th & 3rd & - & - \\
\hline
L2F/INESC-ID         & - & - & - & - & 1st & 1st & 1st & 1st & - & - & - & - \\
\hline
\hline
{\bf Blue Man Group} & 2nd & 2nd & 1st & 1st & 4th & 4th & 3rd & 2nd & 2nd & 3rd & 2nd & 1st \\
\hline
\end{tabular}
\end{table*}

In Table~\ref{tab:competion_results}, we list the results of all methods that we evaluated, considering the
labels of the blind test data made available after the competition.

\begin{table*}[h]
\centering
\caption{Competition results, considering the blind test set.}
\label{tab:competion_results}
\begin{tabular}{|c|c|c|}
\hline
{\bf Configuration} & {\bf Similarity} & {\bf Entailment}	\\
\hline
Baseline: Bag of Words Overall & 0.47 &  \\
\hline
Kenter's features - SVR(M) PT-BR & 0.64 & 81.65/0.52\\
Kenter's features - SVR(M) PT-PT  & 0.64 & 77.60/0.61\\
Kenter's features - SVR(M) Overall & 0.63 & 79.62/0.58 \\
Kenter's features - Lasso PT-BR	& 0.65 & \\
Kenter's features - Lasso PT-PT  & 0.63 & \\
Kenter's features - Lasso Overall & 0.63 & \\
\hline
LSTM várias camadas + reg L2 & & \\ 
LSTM várias camadas + reg L2 + features &  & \\
CNN &  & \\
LSTM várias camadas + reg L2 + features + Full Data &  & \\
CNN + Cos PT-BR &  & \\
LSTM + Cos PT-BR & & \\
LSTM + Cos Overall &  & \\
LSTM + Concat + Kenter's features PT-BR &  & \\
LSTM + Cos + Kenter's features PT-BR &  & \\
LSTM + Dot + Kenter's features PT-BR &  & \\
LSTM (Cos) + BOW (Cos) + Kenter's features PT-BR & & \\
CNN (Cos) + BOW (Cos) + Kenter's features PT-BR &  & \\
BOW (Cos) + BR  & & \\
LSTM (Cos) + CNN (Cos) + BOW (Cos) + Kenter's features PT-BR & & \\
\hline
\end{tabular}
\end{table*}

\section{Conclusions and Future Work}
In this paper we presented the methods and results followed by our team, to participate in the ASSIN 
competition, and evaluate the results obtained compared with the other teams. In our case, we decided
to exploit word vector-based approaches, following two distinct strategies. But given the bad results 
of the second strategy in the evaluation datasets, we pursued in the competition only the method from
the first strategy, based on a state-of-the-art feature set for semantic similarity encoding. With
this approach, we have been ranked best in the entailment recognition task and in semantic similarity
evaluation, achieving the best F1 score overall, and the best accuracy and F1 score in the PT-BR 
dataset. In semantic similarity, our best result was the second place in the PT-BR set.

The experience of participating in the competition has been very valuable, and we expect to continue
working in the problems to improve our method and the results. One future work is to understand better
why siamese networks have not perform as well as the first strategy in these problems. Also, we would
like to better investigate Kenter's features, in order to improve this feature set on these tasks.

\bibliographystyle{fullname_pt}
\bibliography{references.bib}

\end{document}